%% file: main.tex
\pdfoutput=1
\documentclass[11pt]{article}

\usepackage[]{naacl2021}

\usepackage{times}
\usepackage{latexsym}

\usepackage[T1]{fontenc}
\usepackage[utf8]{inputenc}

\usepackage{microtype}

\usepackage[normalem]{ulem}
\usepackage{url}
\usepackage{float}
\usepackage{wrapfig}
\usepackage{multirow}
\usepackage{booktabs}
\usepackage{graphicx}
\usepackage{longtable}
\usepackage{csquotes}
\usepackage{geometry}
\usepackage{amsmath}
\usepackage{caption}
\usepackage{pgfplots}
\pgfplotsset{compat=1.17}
\usepackage{tikz}
\input{tikz}

\showboxdepth=\maxdimen
\showboxbreadth=\maxdimen
\title{Identifying Helpful Sentences in Product Reviews}

\author{Iftah Gamzu\textsuperscript{1} \;\; Hila Gonen\textsuperscript{1} \;\; Gilad Kutiel\textsuperscript{1} \;\; Ran Levy\textsuperscript{1} \;\;  Eugene Agichtein\textsuperscript{2,3}\\
	\texttt{\{iftah,gonenhi,gkutiel,ranlevy,eugeneag\}@amazon.com} \\
	\\ \textsuperscript{1}Amazon, Tel-Aviv, Israel \\
	\textsuperscript{2}Amazon, Seattle, WA, USA \\
    \textsuperscript{3}Emory University, GA, USA}

\begin{document}
	\maketitle
	\begin{abstract}
		\input{abstract}
	\end{abstract}
	
	\section{Introduction}
	\input{intro}

	\section{Related Work}
	\input{related}

	\section{Representative Helpful Sentences}
	\label{define}

\input{task_definition}

	\section{Helpful Sentences Annotation}
	\input{helpful_sentences_dataset}

	\section{Surfacing Representative Helpful Sentences}
	\label{sec:system}
	\input{system}

	\section{Evaluation}\label{sec:eval}
	\input{evaluation}

	\section{Conclusion}
	\input{conclusion}

	\section{Ethical Considerations}

\input{ethics}

	\bibliographystyle{acl_natbib}
	\bibliography{main}

	\appendix

\input{appendix}

\end{document}

%% file: tikz.tex
\usetikzlibrary{
    shapes,
    arrows,
    patterns,
    positioning,
    datavisualization,
}

\tikzstyle{split} = [
    diamond, 
    draw, 
    fill=blue!20, 
    text width=4.5em, 
    text badly centered, 
    node distance=3cm, 
    inner sep=0pt
]

\tikzstyle{block} = [
    rectangle, 
    draw, 
    fill=blue!20, 
    text width=6em, 
    text centered, 
    rounded corners, 
    minimum height=4em
]

\tikzstyle{filter} = [
    block,
    dashed,
    fill=red!20,
]

\tikzstyle{line} = [
    draw, 
    -latex'
]

\tikzstyle{cloud} = [
    draw, 
    ellipse,
    fill=red!20, 
    node distance=3cm,
    minimum height=2em
]

%% file: abstract.tex
In recent years online shopping has gained momentum and became an important venue for customers wishing to save time and simplify their shopping process. A key advantage of shopping online is the ability to read what other customers are saying about products of interest. In this work, we aim to maintain this advantage in situations where extreme brevity is needed, for example, when shopping by voice. We suggest a novel task of extracting a single representative helpful sentence from a set of reviews for a given product. The selected sentence should meet two conditions: first, it should be helpful for a purchase decision and second, the opinion it expresses should be supported by multiple reviewers. This task is closely related to the task of Multi Document Summarization in the product reviews domain but differs in its objective and its level of conciseness. We collect a dataset in English of sentence helpfulness scores via crowd-sourcing and demonstrate its reliability despite the inherent subjectivity involved. Next, we describe a complete model that extracts representative helpful sentences with positive and negative sentiment towards the product and demonstrate that it outperforms several baselines.

%% file: intro.tex
Customer reviews are known to be a valuable source of information for potential buyers. 
This is evident from the high engagement of customers with reviews, for example by up-voting a review for its helpfulness.\footnote{At the time of writing this paper, a review for the Echo Dot 3rd generation received more than 10K votes.} 
As online shopping platforms attract more traffic it is becoming increasingly difficult to consume the wealth of information customers share. 
For this reason, helpful reviews (defined as such by customers) are made more visible than those that are less helpful.

The topic of review helpfulness has attracted a lot of academic interest in which reviews were always considered as a whole (see~\newcite{diaz2018modeling} for a survey). 
However, in some scenarios, such as the limited real-estate in mobile screens, or in voice interactions with a virtual assistant, presenting a full review is impractical and the need to automatically extract helpful excerpts arises. While in the mobile scenario, a persistent customer may still be able to read the entire review, the voice scenario is inherently more challenging as it demands patience and focus from the customer, while the assistant reads the text out loud. As a result, the need for extreme brevity and the ability to understand what matters most to customers becomes crucial.

In addition to brevity and helpfulness, another desirable property from the extracted content is being faithful to the reviews as a whole. Indeed, a customer looking for relevant and helpful reviews, often interacts with more than one review before making their decision, trying to pinpoint those helpful bits of information that are shared by multiple reviewers. 
This process is tedious because of the sheer amount of reviews and biased because of the order they appear in. A system that aims to replace this process while maintaining trust in the content it provides should be able to extract concise helpful texts that repeat across multiple reviews, indicating that they are faithful to the reviews' content (from here onward we shall refer to such texts as ``faithful'').

Our goal is to extract such sentences, i.e., sentences that are both \textbf{helpful} for a purchase decision and \textbf{faithful}. To this end, we first define two new notions: A \textit{Helpful Sentence} is a sentence which is considered helpful by the average customer in their purchase decision process.  A \textit{Representative Helpful Sentence (RHS)} is a helpful sentence that is also highly supported, that is, the ideas it expresses appear in multiple reviews for the given product (not necessarily in the exact same wording). 

It is traditionally assumed that judging the importance of a text excerpt requires reading the entire text. We challenge this assumption, at least in the domain of product reviews, and collect a dataset of single review sentences with their helpfulness scores by averaging the scores assigned to them by multiple crowd workers. We show that despite the highly subjective nature of this task, and despite the fact that workers are exposed to sentences without their surrounding context, the resulting scores are reliable. Using the data we collected, from 6 different categories, ranging from Electronics to Books, we train and evaluate several supervised algorithms to predict helpfulness score, which achieve promising results. Finally, we present an initial implementation of a model that given a set of product reviews, extracts a single positive RHS (supports the purchase) and a single negative RHS (opposes the purchase).

In summary, the main contributions of this work are:
(1) We propose a novel task that given a set of reviews for a product, outputs a single sentence that is both helpful for a purchase decision and supported by multiple reviewers; (2) We show that the helpfulness of a sentence can be reliably rated based solely on the sentence, allowing for an efficient dataset creation. These helpfulness scores can be leveraged for other tasks such as highlighting important parts of a review; (3) We publish a novel dataset of sentences taken from customer reviews along with their helpfulness score;\footnote{The dataset is available at \url{https://registry.opendata.aws/helpful-sentences-from-reviews/}.} (4) We develop an end-to-end model for our task that shows promising results and outperforms several baselines.

%% file: related.tex
\paragraph{Review Helpfulness Modeling and Prediction}
Customer reviews are a valuable source of information for customers researching a product before making a purchase \cite{Zhu2010ImpactOO}. \newcite{diaz2018modeling} survey recent work on the tasks of modeling and predicting review helpfulness. While some researchers treat helpfulness votes as ground-truth, others have argued that these votes are not good indicators for actual review helpfulness \cite{liu-etal-2007-low,Tsur2009RevRankAF,yang-etal-2015-semantic}. 

Some general observations have been made based on helpfulness votes, e.g., review length has been shown to be strongly correlated to helpfulness \cite{KimEtAl2006,liu-etal-2007-low,Otterbacher2009,MudambiAndSchuff2010,PanAndZhang2011,yang-etal-2015-semantic}. Another widely-agreed indication for review helpfulness is the review star rating \cite{KimEtAl2006,MudambiAndSchuff2010,PanAndZhang2011}.

A related dataset was presented in \newcite{almagrabi2018corpus}. The main advantages of the dataset we create over this previously suggested one are: (1) \textbf{Binary vs. continuous scores} -- We use continuous scores rather than binary scores. Our aim is to surface the most helpful sentences, which is not possible if many of the sentences are annotated as equally helpful; (2) \textbf{Range of products/domains} -- The previous dataset includes only 5 products, all from the Electronics domain. Our dataset is significantly more diverse, providing annotations for 123 products from 6 different domains, allowing to evaluate a model's ability to generalize across domains.

\paragraph{Product Review Summarization}

The most common approach for product review summarization, which centers the summary around a set of extracted aspects and their respective sentiment, is termed \emph{aspect based summarization}.
One of the early abstractive works, by \newcite{hu2004mining}, was designed to output lists of aspects and sentiments. Other works target a traditional summarization output and at times somewhat simplify the task by assuming aspects or seed words are provided as input \cite{Gerani2014prodReviewAbsSumm,angelidis2018oposum,yu2016reviewSumm}. 
Recently advances were made on unsupervised abstractive reviews summarization, by leveraging neural networks \cite{chu2019meansum,bravzinskas2020unsupervised} followed by a few shot variant \cite{bravzinskas2020few}.

Extractive summarization include earlier works such as \newcite{Carenini2006MultiDocumentSO,Lerman2009SentimentSE} and \newcite{XiongAndLitman2014} who suggested to use review helpfulness votes as means to improve the content extraction process. More recently, \newcite{Tan2017SentenceRW} suggested a novel generative topic aspect sentiment model.

%% file: task_definition.tex
\paragraph{Task Definition}
\label{para:def}
In this work, we focus on summarization of reviews in the setting of shopping over voice with the help of a virtual assistant. 
Our goal is to provide users with content that is both \textbf{helpful} and \textbf{faithful} in this challenging setting where the information the user can absorb is extremely limited. First, we aim to maximize the informativeness, while maintaining brevity. To this end, we introduce a new notion of \textit{helpful sentences} -- sentences which the average customer will consider as helpful for making a purchase decision. Next, to ensure faithfulness, we introduce the notion of \textit{support} for a given sentence -- the number of review sentences with a highly similar content. We seek to \textbf{automatically} identify a helpful sentence with a wide support, which we term \textit{representative helpful sentence (RHS)}. Note that Representative Helpful Sentences, being supported by many similar sentences, are by construction faithful to the review pool from which they are extracted. We restrict ourselves to single sentences that are extracted as-is from product reviews, as this serves as another mechanism to ensure faithfulness. We do not restrict the number of reviews in the input.
Table~\ref{tbl:helpexamples} presents a few helpful sentences for example, as extracted by our model (see Section~\ref{sec:system}).

\begin{table}[ht]
	\centering
	\resizebox{\columnwidth}{!}{
		\begin{tabular}{p{1.6cm}p{7cm}} \toprule
			Product & Representative Helpful Sentence \\
			\midrule
			 Toy Pirate Ship & It was easy to put together, is the perfect height, and very durable.  \\
			\midrule
			 Headphones &They fit well and produce good sound quality. \\
			\midrule
			Speakers &Quality good, price okay, sound output great. \\
			\bottomrule        
	\end{tabular}}

	\caption{\label{tbl:helpexamples}Example Representative Helpful Sentences.}
\end{table}

Our task resembles the well known (extractive) customer review summarization task \cite{hu2004mining} but differs in several important aspects. First, its output is very concise due to the extreme space constraint, resembling the extreme summarization task \cite{narayan2018don}, which however, deals with news articles and outputs an abstractive summary. In our application there is low tolerance for factually incorrect summaries, so we choose extraction over abstraction. Second, we do not restrict the system's output to aspect based opinions, as we find that sometimes factual content may also be quite helpful. Third, while traditional summarization systems favor information that appears frequently in the source documents, we target information that is both frequent and helpful.

\paragraph{Subjectivity}

As mentioned above, review helpfulness scores are derived from votes of actual customers. Deciding on whether or not to up-vote a review is a subjective decision as different customers may value different product qualities. However, the underlying assumption of the voting mechanism is that reviews with many up-votes are indeed helpful for the average customer. Restricting the user to a single sentence makes matters even more challenging as it cannot possibly discuss all the product merits and shortcomings. To emphasize the subjectivity involved in assigning a helpfulness score for a standalone sentence, consider the examples in Table~\ref{tbl:examples}. The first example may be helpful for parents looking to buy a book for their children but entirely unhelpful for adults who wish to purchase the book for themselves. Similarly, the second one is more helpful to readers of extreme height (short or tall) than to those of medium height. 

\begin{table}[ht]
	\centering
	\resizebox{\columnwidth}{!}{
	\begin{tabular}{p{1.2cm}p{7.5cm}} \toprule
		Product & Sentence \\
		\midrule
		Harry Potter book & Finding 1 book that keeps your child intrigued and helps him or her develop a love for reading is amazing. \\
		\midrule
		Jump rope & It's a pretty standard jump rope but it's really nice and you can adjust the length which is perfect because I'm really short. \\
		\bottomrule        
	\end{tabular}}
	\caption{\label{tbl:examples}Review sentence examples.}
\end{table}

Despite the evident subjectivity, we assume that there exists an ``average helpfulness'' score for every sentence, which can be estimated by averaging the ratings of multiple crowd workers. In the following section we establish this assumption by compiling a new dataset of sentences along with their helpfulness scores, and showing quantitatively that the annotations in our dataset are consistent and reliable.

%% file: helpful_sentences_dataset.tex
Our main contribution in this work lies in the notion of \textit{helpful sentences} and the ability to identify such sentences without observing entire reviews. In what follows, we describe the process of compiling a dataset of sentences along with their helpfulness scores using crowdsourcing. Note that this dataset is intended solely for scoring helpfulness of sentences. Faithfulness is ensured by other means which are not reflected in the dataset, i.e. by requiring a RHS to have a wide support of similar sentences, as discussed in section~\ref{para:def} and implemented in our model, as described in Section~\ref{sec:model}.

\subsection{Annotation Task}
\label{ssec:data}

We consider a subset of 123 products arbitrarily selected from the Amazon.com website, so that each has at least 100 customer reviews and they (approximately) equally represent 6 different categories (Toys, Books, Movies, Music, Camera and Electronics).
We started with 45,091 reviews, split them into 210,121 sentences and randomly selected a train set with 20,000 sentences, and a test set with 2,000 sentences.
We asked annotators to rate each sentence according to how helpful it is for reaching a purchase decision, using the Appen platform.\footnote{\url{www.appen.com}} Ratings were provided on a 3-level scale of \textbf{Not Helpful} (0), \textbf{Somewhat Helpful} (1), or \textbf{Very Helpful} (2). The final helpfulness score of a given sentence was set to the average rating. See Section~\ref{app:task} in the Appendix for more details on the annotation task guidelines.

Each example was rated by 10 different annotators in the training set and 30 different annotators in the test set. Initial experiments revealed that 10 annotations per sentence, while noisy, are still sufficient to train a model.  We observed that the number of annotators used to calculate the test set affects the evaluation. This is due to the subjective nature of this task and the observed helpfulness score that becomes closer to its ``real'' score as the number of votes collected for each sentence increases. Table~\ref{tbl:test-curve} demonstrates the effect the number of annotators used to rate each example in the test set has on the final evaluation. It shows that after fixing the model and predictions, the evaluation score (Pearson correlation in this case) increases as we average more votes. From our experience, there is no gain beyond 30 votes per sentence for this particular task.
\begin{table}[ht]
	\centering
	\resizebox{\columnwidth}{!}{
		\begin{tabular}{llllll} 
			\toprule 
			\# of votes  & 1     & 10    & 20    & 25    & 30    \\
			Pearson & 0.523 & 0.776 & 0.822 & 0.831 & 0.838 \\
			\bottomrule
	\end{tabular}}
	\caption{For a fixed prediction the correlation between the prediction and the scores obtained by averaging individual scores increases as we consider more votes per sentence. The phenomenon is not unique to correlation.\label{tbl:test-curve}}
\end{table}

We observe a skewed helpfulness distribution with a fairly high mode of $1.3$ which shows that the raters did not provide random answers. Furthermore, under the assumption that most review authors aim for their reviews to be helpful, we should expect a distribution that is skewed towards higher scores. See Section~\ref{app:task} in the Appendix for a depiction of the helpfulness distribution within the train set.

Table~\ref{tab:top-bottom-sentences} presents the most helpful sentence, a sentence that is somewhat helpful (with a median score) and the least helpful sentence from the test set for particular headphones as perceived by the annotators.

\begin{table}[ht]
	\centering
	\resizebox{\columnwidth}{!}{
		\begin{tabular}{@{}p{0.9\linewidth}c@{}} \toprule
			Sentence & Helpfulness \\
			\midrule
			Really great headphones, especially for \$25, but honestly, they sound better than my gaming headset and my DJ headphones in many respects. & 1.97 \\
			\midrule
			Call quality just can't be beat for ear buds & 1.47 \\
			\midrule
			Any thoughts from others? & 0 \\
			\bottomrule        
	\end{tabular}}
	\caption{The most helpful, a somewhat helpful and the least helpful sentences for particular headphones.\label{tab:top-bottom-sentences}}
\end{table}

\subsection{Annotation Analysis}
As mentioned earlier, rating sentence helpfulness is a highly subjective task, and some disagreement is expected. Nevertheless, we argue that the data we collected is reliable and demonstrate it through the three following experiments. 

\paragraph{Inter-annotator Agreement} We compute agreement in the spirit of the analysis performed in~\cite{snow2008}. For each annotator, we restrict the data to the set of rows that they completed and compute the Pearson correlation between their answers against the average of all other annotators. Finally, we take the average across all annotators after removing the worst $10\%$ annotators according to the method of~\cite{DawidAndSkene}.
We get an average of $0.44\pm0.01$ Pearson correlation on the train set (10 annotators per row) and $0.57\pm0.02$ on the test set (30 annotators per row), which demonstrates good agreement given the subjective nature of this task.\footnote{This scores are comparable, for example, with the scores reported in~\newcite{snow2008} for the highly subjective Affective Text Analysis task.}
We also randomly split the annotators into two disjoint sets and calculated the correlation
between the corresponding scores.
There was a correlation of $0.49$ for the train set and $0.81$ for the 
test set. 

\paragraph{Internal Consistency}
A necessary condition for ensuring reliability is that similar sentences get similar helpfulness scores. We verify that our crowd-sourced test data meets this requirement by measuring the standard deviation of the helpfulness scores within groups of similar sentences. We use the sentence-transformers embeddings of \newcite{reimers-2019-sentence-bert} which were optimized for computing semantic similarity. 
For each sentence in the test set, we construct its semantic neighborhood by grouping together all sentences with high similarity. For each non-singleton group, we measure the standard deviation of the helpfulness score and compare it with the standard deviation of a similarly sized group of random sentences from the test set. We expect to get a tighter distribution of helpfulness scores within the similarity groups (compared to the random groups) if the data is internally consistent. Indeed, we found $217$ groups with an average standard deviation of $0.16$ while the average standard deviation of the corresponding random groups was $0.29$.\footnote{The differences were statistically significant with a p-value of $7E-20$ using a paired two-tailed t-test.}

\paragraph{Sentence Helpfulness vs. Review Helpfulness}
As the third and final reliability analysis, we compare the crowd helpfulness scores with review helpfulness votes taken from the Amazon.com website. We consider reviews for the $123$ products selected earlier, and extract two subsets. The first (the helpful set) is the
set of all reviews with at least $50$ helpful votes. The second (the unhelpful set) is the set of all reviews with no helpful votes. See Section~\ref{app:analysis_annotation} in the Appendix for statistics on the two subsets.
We randomly select $500$ sentences from each set and collect crowd helpfulness ratings. For each set we calculate the mean helpfulness score and the ratio of sentences with helpfulness score greater than 1 and 1.5 respectively.
Table~\ref{tab:contrast-result} shows the results which demonstrate a higher mean helpfulness score in the helpful set.\footnote{The difference is statistically significant with a p-value of approximately $0.0079$ using a t-test with an equal variance assumption as well as t-test with different variance assumption, a.k.a Welch’s t-test.}

\begin{table}[ht]
	\centering
	\resizebox{0.9\columnwidth}{!}{
	\begin{tabular}{@{}lccc@{}} \toprule
		& Mean & Ratio & Ratio \\
		& Score & Score $>1$ & Score $>1.5$\\
		\cmidrule{2-4}
		Helpful Set     & 1.21  & 70\%  & 18\% \\
		Unhelpful Set   & 1.15  & 64\%  & 14\% \\
		\bottomrule
	\end{tabular}}
	\caption{
		Contrasting Sentence Helpfulness Score with Review Helpfulness votes --- Results.%
		\label{tab:contrast-result}}
\end{table}

These results indicate that helpful reviews tend to include more helpful sentences on average. However, as can be expected, the differences are not dramatic. Looking at the average length of reviews sheds some more light on the differences: a helpful review is almost 10 times longer than a non helpful review on average. This means that in order for a review to be helpful it must provide details, a requirement that a single sentence simply cannot meet. Therefore, we conjecture that a helpful sentence captures the most essential statements made in the review while a helpful review is one that includes details and justifies its rating.

\subsection{Analysis of Helpful Sentences}
A brief examination of the crowd-sourced data reveals two sentence characteristics that contribute to the helpfulness of a sentence: the length of the sentence and the sentiment, which is more strongly correlated with helpfulness.

\paragraph {Length} The Pearson correlation between the length (in characters) and the helpfulness score on the test set is $0.37$. This correlation is expected, since longer sentences can potentially convey more information and thus tend to be more helpful.

\paragraph {Sentiment} We use Amazon AWS comprehend sentiment analysis tool\footnote{\url{https://aws.amazon.com/comprehend/}} to classify each sentence into one of four sentiment classes: positive, negative, neutral and mixed. We got a negative Pearson correlation of $-0.53$ between the helpfulness scores of the sentences and the scores assigned to the neutral class. To better understand this relationship, we define a helpful sentence as one with score greater or equal to $1.5$ and a sentence with sentiment as one that is not in the neutral class, and estimate two conditional probabilities:
\begin{align}
P(\text{Helpful}\mid\text{Sentiment}) = 0.15 \nonumber \\ 
P(\text{Sentiment}\mid\text{Helpful}) = 0.68 \nonumber
\end{align}
This shows that having sentiment is an important condition for a sentence to be helpful, but it is not a sufficient condition. We indeed observed that sentences with sentiment that do not provide additional reasoning or details do not get high helpfulness scores. Some related examples from reviews can be found in Section~\ref{app:analysis_help} in the Appendix.

%% file: system.tex
\label{sec:model}

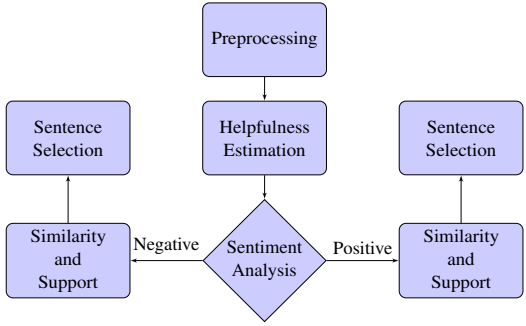
\begin{figure}[ht]
\centering
    \resizebox{0.9\columnwidth}{!}{
        \begin{tikzpicture}[node distance=1cm, auto]
            \node[block] (token) {Preprocessing};
            \node[block, below=5mm of token] (help) {Helpfulness Estimation};
            \node[split, below=5mm of help] (sentiment) {Sentiment Analysis};
            
            \node[block, right=1.5cm of sentiment] (pos sim) {Similarity \\ and \\ Support};
            \node[block, above=of pos sim] (pos sentence) {Sentence Selection};
            
            \node[block, left=1.5cm of sentiment] (neg sim) {Similarity \\ and \\ Support};
            \node[block, above=of neg sim] (neg sentence) {Sentence Selection};
            
            \draw[line] (token) to (help);
            \draw[line] (help) to (sentiment);
            
            \draw[line] (sentiment) -- node [above, sloped]{Negative} (neg sim);
            \draw[line] (neg sim) to (neg sentence);
            \draw[line] (sentiment) -- node [above, sloped]{Positive} (pos sim);
            \draw[line] (pos sim) to (pos sentence);
        \end{tikzpicture}
    }
    \caption{High-level overview of our model.\label{fig:sys}}
\end{figure}

We now turn to create an end-to-end model for surfacing representative helpful sentences (RHS): given a set of reviews for a certain product, we aim to output a single RHS with positive sentiment and a single RHS with negative sentiment. Figure~\ref{fig:sys} depicts the different sub-components of our model. Given a set of reviews, we preprocess the input and predict helpfulness scores for each of the sentences. Next, we analyze the sentiment of each sentence and separate into positive and negative sets. Following that, the support of each sentence is determined, and finally we select the RHS sentence based on its helpfulness score and its support. In what follows, we describe each of the components in details.

\paragraph{Preprocessing} 
We remove HTML tags and split the cleaned reviews into sentences. The sentences are then filtered by removing sentences of extreme length (both short and long). See Section~\ref{app:surface} in the Appendix for additional details.

\paragraph{Helpfulness Estimation}
This component assigns a helpfulness score for each sentence and removes all sentences with score below $1$. This filtering serves two purposes: First, it ensures that we do not output any sentence in case there is no helpful sentence in the product reviews. Second, it reduces the runtime of the downstream Similarity and Support component which is quadratic in the number of sentences. 

We experiment with three helpfulness models and find that a pre-trained BERT~\cite{devlin2018bert} fine-tuned on our training data performs best. The two other models we compare are: (1) \textsc{TF-IDF}: a model that treats each sentence as a bag-of-words. We use \texttt{TfidfVectorizer} from the \texttt{sklearn} package to convert each sentence into a vector and then fit a Ridge regression model on top of it; (2) \textsc{ST-Ridge}: a model that fits a Ridge regression on top of the Sentence-Transformers embedding \cite{reimers-2019-sentence-bert}.

We use 3 measures for evaluation: Mean Squared Error (MSE), which is the traditional measure for regression, Pearson correlation between the predicted score and the ground-truth score, and finally a ranking measure that evaluates the quality of the top ranked sentence (NDCG@1). 
The results are depicted in~Table~\ref{tab:eval}. The \textsc{TF-IDF} model has an acceptable performance but it suffers from out-of-vocabulary problem and ignores the sentence as a whole, for example, the model predicts a higher score than that of the annotators to the sentence ``fantastic brilliant amazing superb good''. In order to gain some understanding into what constitutes a helpful sentence, we checked the top positive and negative features of this model.\footnote{Top-10 positive features: great, sound, quality, good, excellent, price, easy, lens, recommend, perfect. Top-10 negative features: bought, review, know, don, got, amazon, gift, reviews, christmas, order.} We observed that the top positive words include sentiment words and product aspects. The results, however, indicate that these features are not sufficient to evaluate the helpfulness in a more fine-grained manner. The \textsc{ST-Ridge} model significantly outperforms the \textsc{TF-IDF} model in all metrics. Finally, the BERT model is significantly better than the \textsc{ST-Ridge} model in terms of MSE and Pearson correlation.

\begin{table}[ht]
	\centering
	\resizebox{\columnwidth}{!}{
		\begin{tabular}{llll} 
			\toprule 
			& MSE & Pearson & NDCG@1 \\
			\cmidrule{2-4} 
			\textsc{Random} & $0.5\pm0.026$ & $0.018\pm0.065$ & $0.68\pm0.044$ \\
			\textsc{TF-IDF} & $0.09\pm0.006$ & $0.63\pm0.055$ & $0.91\pm0.022$ \\
			\textsc{ST-RIDGE} & $0.062\pm0.0042$ & $0.78\pm0.037$ & $0.94\pm0.015$ \\
			\textsc{BERT} & $0.053\pm0.0037$ & $0.84\pm0.022$ & $0.95\pm0.015$ \\
			\bottomrule
	\end{tabular}}
	\caption{Evaluation of Helpfulness Prediction (with confidence intervals).\label{tab:eval}}
\end{table}

\paragraph{Sentiment Analysis}
In this step, we employ the Amazon AWS comprehend sentiment analysis tool to assign each sentence a sentiment class and a score for each of the four classes: positive, negative, neutral and mixed. Sentences with a neutral or mixed classes are removed and all the rest are divided into a positive set and a negative set. The purpose of this step is twofold: first, the separation allows us to output a final sentence for both positive and negative sentiments. Second, we gain more confidence that semantically similar sentences (as measured in the downstream Similarity and Support component) have indeed the same meaning (and not the exact opposite).

\paragraph{Similarity and Support}

At this stage we aim to compute the support of each sentence, which we define as the size of the set of highly similar sentences. Formally, for a given sentence $s_i$, its support is $|\{s_{j \neq i}|sim(s_i,s_j)>\sigma\}|$, where $\sigma$ is a predefined threshold.

To compute the similarity, we convert each sentence pair to the corresponding representations and compute the cosine similarity. In order to get the most accurate results, we compare several sentence representations on the semantic similarity task: the Sentence Transformers~\cite{reimers-2019-sentence-bert}, 
the Universal Sentence Encoder (USE)~\cite{use},
FastText~\cite{mikolov2018advances}, 
and a bag-of-words representation weighted by the inverse document frequency. We find that the Sentence Transformers embeddings perform best. 

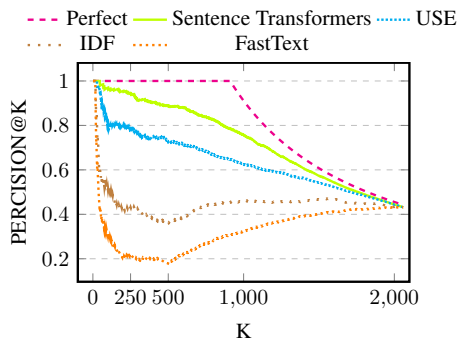
\begin{figure}[ht]
	\centering
	\resizebox{0.8\columnwidth}{!}{
		\begin{tikzpicture}
			\begin{axis}[
				width=0.97\linewidth,
				height=5.5cm,
				legend style={at={(0.5,1)},anchor=south,draw=none},
				legend entries={Perfect,Sentence Transformers,USE,IDF,FastText},
				legend columns={3},
				grid style=dashed,            
				no markers,
				very thick,
				xmin=-100,
				xmax=2100,
				xtick={0,250,500,1000,2000},
				xlabel=K,
				ymajorgrids=true,
				ylabel=PERCISION@K,
				]
				\addplot[magenta, dashed] table [col sep=comma, x=ID, y=perfect] {data/similarity.csv};
				\addplot[lime] table [col sep=comma, x=ID, y=bert] {data/similarity.csv};
				\addplot[cyan, densely dotted] table [col sep=comma, x=ID, y=use] {data/similarity.csv};
				\addplot[brown, loosely dotted] table [col sep=comma, x=ID, y=idf] {data/similarity.csv};
				\addplot[orange, dotted] table [col sep=comma, x=ID, y=ft] {data/similarity.csv};
			\end{axis}
	\end{tikzpicture}}
	\caption{Comparison of Similarity Measures.}
	\label{fig:sim}
\end{figure}

To compare the methods, we sample 300,000 sentence pairs from the reviews of our 123 products, compute the similarity scores on this sample and select the top 500 pairs using each of the methods. We next consider the union of the above pairs to form a dataset of 2,035 pairs. We ask human annotators to determine if the sentences of each pair have a roughly similar meaning or not. We then calculate the precision at K (for K between 1 and 2,035) for each of the methods.
As can be seen from Figure~\ref{fig:sim}, Sentence-Transformers is superior to the other methods.

Finally, we derived a precision-oriented similarity score threshold ($\sigma=0.876$) for Sentence Transformers that achieves a precision of $0.9 \pm 0.286$ and a recall of $0.46 \pm 0.022$ where the recall is estimated based on the set of 2,035 pairs.

\paragraph{Sentence Selection}
The Sentence Selection component is in charge of selecting a single sentence that is both helpful and well supported. We enforce a minimum support of $5$, as we observed that such a limit increases the overall quality of the sentence and avoids surfacing esoteric opinions.
After applying this threshold, we rank the remaining sentences according to the formula: $\text{support} \times{} \text{helpful}^\alpha$,
where $\alpha$ is a boosting parameter. 
To derive an appropriate value for $\alpha$ we conducted another annotation task and obtained a value of $\alpha=38.8$ that gives a lot of emphasis to the helpfulness score. We describe this in detail in Section~\ref{app:surface} in the Appendix.

%% file: evaluation.tex
The evaluation of our end-to-end model is challenging and does not have a natural scheme.
Recall that we do not restrict our input to small random samples of  the review set, as commonly done in review summarization, and was shown to produce biased results \cite{shapira2020massive}. Instead, we allow for dozens or hundreds of reviews per product. Thus, we cannot expect annotators to carefully read the full input before choosing an RHS. Nonetheless, we show that our notion of helpfulness is indeed useful for surfacing important review content by comparing our models to previous summarization works in two different settings.

\paragraph{Single Review Summarization}

In this evaluation we only consider the helpfulness component, as means to create an extractive summary comprised of a single sentence.

Abstractive single review summarizers \cite{MaEtAl2018,IsonumaEtAl2019,wang2018self} are not suitable for comparison as these works are trained on header-like summaries of 4.36 words on average, much shorter than our extractive, one-sentence output.
Instead, we consider the unsupervised single document summarization algorithm Textrank\footnote{We used the implementation of \url{https://pypi.org/project/sumy/}} \cite{mihalcea2004textrank}.
Textrank, which is extractive and can output any number of sentences, is a viable candidate for comparison as our goal is not to achieve SOTA results on this task, but rather to demonstrate that the helpfulness model can produce good extractive summaries without being trained on reference summaries.

We selected a sample of $300$ reviews, in which the prediction of the two algorithms differed (the output was exactly the same on $28\%$ of the reviews), and asked crowd workers to rate each of the selected sentences in a 5-level scale according to how helpful the selected sentence was for a purchase decision (our objective) and according to how well the selected sentence summarized the review (the traditional objective). Each sentence was annotated by $5$ workers, where the sentences of the two algorithms appeared one next to the other but in random order. Table \ref{tab:sds} summarizes the results, showing our method is superior in both aspects.\footnote{The results are statistically significant using a 1-tail paired t-test, with a p-value of $1.05E-06$ for helpfulness and $0.005$ for summarization.}

\begin{table}[ht]
	\centering
	\resizebox{\columnwidth}{!}{
	\begin{tabular}{@{}lllll@{}} \toprule
		& Helpfulness & & Summarization &\\
		& Mean & Std & Mean & Std \\
		Helpful Sentence & $3.41$ & $1.11$ & $3.34$ & $1.05$ \\
		Textrank & $3.28$ & $1.16$ & $3.27$ & $1.10$ \\   
		\bottomrule
	\end{tabular}}
	\caption{Single Document Summarization - comparison to TextRank.\label{tab:sds}}
\end{table}

\paragraph{End-To-End Evaluation}

Our complete model resembles the task of Multi-Document Summarization (MDS) which ideally consumes the entire set of reviews related to a specific product and outputs a single summary or a single sentence in our case. In practice, MDS is applied to document sets of relatively small sizes, which significantly reduces the potential impact of our Similarity and Support sub-component. In order to put our evaluation in context of prior work, we evaluate our model with two minor modifications tailored for small review sets: we relax the similarity threshold to 0.75 and remove the minimal support constraint.  We only consider the positive sentences in this evaluation, as the majority of the reviews are positive.

We use the dataset published in~\newcite{bravzinskas2020unsupervised} which covers $60$ products from 4 different categories (Cloth, Electronics, Health Personal Care and Home Kitchen)\footnote{4 of the products in this dataset are no longer available on amazon.com and we omitted them from the evaluation.} of which only $1$ category is included in our own data. Each product has $8$ reviews and $3$ reference summaries written by humans.
We evaluate our model in a straight forward manner by comparing the sentences selected by our model to sentence rankings provided by humans.

We ask expert annotators (one annotator per example) to read the reviews and rate each sentence from the reviews on a scale of 1 to 5.
A score of 1 means that the sentence does not help to make a purchase decision or it does not reflect the overall theme of the reviews where a score of 5 means that the sentence is both helpful and aligns well with the common opinions expressed in the reviews.
The mean score of the top sentence for each product is $4.31$, which means that even for products with only $8$ reviews it is common to find a sentence that is both helpful and supported by the reviews.
We evaluate our model by averaging NDCG@K over all products for $K \in \{1, 10\}$.
We compare the performance of our model with two baselines: ranking the sentences in a random order and from the longest to the shortest. Our method outperforms the baselines by a large margin, see Table~\ref{tbl:e2endcg}.

\begin{table}[ht]
	\centering
	\resizebox{0.7\columnwidth}{!}{
	\begin{tabular}{lll} 
		\toprule 
		                            & K=1	& K=10 	\\ 
		\midrule
		Our	Model           & 0.87	& 0.94	\\
		From Longest to Shortest	& 0.60	& 0.68	\\		Random		                & 0.54	& 0.62	\\

		\bottomrule        
	\end{tabular}}
	\caption{Mean NDCG@K score.\label{tbl:e2endcg}}
\end{table}

For the sake of completeness we also report the common MDS evaluation metric, ROUGE~\cite{lin2004rouge}, which does not fully suit our setting, as it is based on n-gram comparisons between the output and golden summaries written by humans, which are typically much longer than a single sentence. In Table~\ref{tbl:rouge} we compare the ROUGE scores of 3 sentence selection variants: our model, a random sentence and an Oracle, i.e., the sentence that maximizes the ROUGE-L score. We also report the results of  Copycat~\cite{bravzinskas2020unsupervised},\footnote{Results are based on our own computation using \url{https://pypi.org/project/py-rouge/}} a state-of-the-art review MDS model. We note that Copycat is not truly comparable to our model due to the significantly different summary length requirement (in this dataset an average sentence contains 74 characters while an average reference summary contains 293 characters). Note, however, that in terms of precision, which is what we aim for with such an extreme ``summary'', the RHS is almost as good as the Oracle and much better than Copycat.

\begin{table}[ht]
	\centering
	\resizebox{0.8\columnwidth}{!}{
	\begin{tabular}{lllll} 
		\toprule 
		% & \multicolumn{3}{c}{Sentence Level} & Summarizer	\\
		& Random & RHS	& Oracle & Copycat  				\\ 
		\cmidrule(lr){2-4}\cmidrule(lr){5-5}
		Rouge-1 f   & 0.127  & 0.166 &  0.250 &   0.297    \\     
		Rouge-1 p   & 0.329  & 0.420 &  0.440 &   0.247    \\     
		Rouge-1 r   & 0.084  & 0.109 &  0.185 &   0.386    \\     
		Rouge-2 f & 0.014 & 0.028 & 0.054 & 0.055 \\
		Rouge-2 p & 0.045 & 0.084 & 0.106 & 0.045 \\
		Rouge-2 r & 0.008 & 0.018 & 0.039 & 0.073 \\ 
		Rouge-L f   & 0.094  & 0.120 &  0.177 &   0.192    \\     
		Rouge-L p   & 0.251  & 0.309 &  0.320 &   0.160    \\     
		Rouge-L r   & 0.061  & 0.078 &  0.129 &   0.249    \\    
 \bottomrule        
	\end{tabular}}
	\caption{ROUGE scores.\label{tbl:rouge}}
\end{table}

\paragraph{Examples of RHS}
We pick two examples from~\newcite{bravzinskas2020few}, depicted in Table~\ref{tbl:eg}, and use our model to extract a single sentence for each.
Each of the examples consists of 8 reviews and a reference summary written by a human.\footnote{We only show the summaries, the complete set of reviews are available in~\newcite{bravzinskas2020few}.} 

The first example deals with Yelp reviews and our model seems to do well even though it was only trained on product reviews from Amazon.  While our extracted sentence is less elaborative compared to the human and abstractive summaries, it gives enough information to make a decision. Note also, that the abstractive summary does not refer to the high pricing. As for the second example, while not covering all aspects of the product, the helpful sentence is faithful to the reviews and aligns with the overall sentiment. The summarizer, on the other hand, contradicts the reviews with regarding to the sandals size.

Recall that these examples are constructed from 8 reviews only, while our model benefits considerably from large number of reviews, which is often the case for popular products. This is due to the greater sentence variety it can choose from and the fact that the support becomes more meaningful as more reviews are available. See Section~\ref{app:stats} in the Appendix for additional examples and some statistics of our model outputs.

\begin{table}[]
	\centering
	\resizebox{\linewidth}{!}{
	\begin{tabular}{p{1.8cm}p{9cm}} 
		\toprule
		\multicolumn{2}{l}{Reviews from Yelp!}  \\
		\midrule
		RHS & They have ridiculous price also bad customer services. \\
		\midrule
		Human & Bennett Medical has poor customer service. Phone calls can take a long time to get answered and leaving voice mails tend to be fruitless. The products are overpriced and take a long time to be refilled. Using this medical supply company can be a hassle. \\
		\midrule
		Summarizer & This medical supply is the worst medical supply company in the valley. The customer service is horrible, the staff is rude, the wait times are long, and the service reps are not helpful at all. Do not recommend this company to anyone. \\
		\midrule[1.5pt]
		\multicolumn{2}{l}{Reviews from Amazon.com}  \\
		\midrule
		RHS & These thongs are fun, festive, flexible and surprisingly comfortable. \\
		\midrule
		Human & These are a very comfortable and cute sandal. This thong sandal goes with a variety of outfits and the cushy sole allows for all day comfort. However, they do run a little small, sizing up provides a better fit. Overall, a reasonably priced shoe that will last for years to come. \\
		\midrule
		Summarizer & These sandals are very cute and comfortable. They fit true to size and are very comfortable to wear. They look great with a variety of outfits and can be dressed up or down depending on the occasion. \\
		\bottomrule        
	\end{tabular}}
	\caption{\label{tbl:eg}Helpful Sentences vs. Abstractive Summarization.}
\end{table}

%% file: conclusion.tex
In this paper we address the challenge of summarizing product reviews with limited space, like when using a virtual assistant. We define a new notion that fits the needs of this setting, a \textit{representative helpful sentence}, and propose a new task accordingly: given a set of product reviews, extract a sentence that is both helpful for a purchase decision and well supported by the opinions expressed in the reviews.

As a first step, we collect and annotate a new dataset of review sentences with their helpfulness scores, and make this dataset available to facilitate further research. Next, we develop an end-to-end model for surfacing representative helpful sentences. Our model combines several necessary components which are optimized for our goal. In order to get a feeling for the performance of our model, we compare our results to summarization tasks that are similar in nature, and show that our model performs better in the aspects we target.

%% file: ethics.tex
In this work, we make use of customer reviews published on Amazon.com. The reviews must comply with Amazon’s Community Guidelines\footnote{\url{https://www.amazon.com/gp/help/customer/display.html?nodeId=GLHXEX85MENUE4XF}} which prohibit offensive, infringing, or illegal content. Amazon encourages anyone who suspects that content manipulation is taking place or that its Guidelines are being violated to notify Amazon. Amazon investigates concerns thoroughly and takes appropriate actions, including removal of reviews that violate these Guidelines, including reviews that contain hatred or intolerance for people on the basis of race, ethnicity, nationality, gender or gender identity, religion, sexual orientation, age, or disability. Among other things, Amazon has a broad license to use, reproduce, publish, and create derivative works from the customer reviews on Amazon.com. The authors of this paper are employees of Amazon and are authorized to use customer reviews in this work.

A small sample of annotated review sentences is released for research purposes according to the provided license.\footnote{\url{https://cdla.dev/sharing-1-0/}} Annotations were conducted by a service provider pursuant to a Service Agreement with Amazon. Under that Service Agreement, the service provider represents and warrants that it complies with all applicable laws, regulations, and ordinances when performing those services.

%% file: appendix.tex
\section{Annotation Task}
\label{app:task}

\begin{figure}[ht]
	\centering
	\includegraphics[width=1\linewidth]{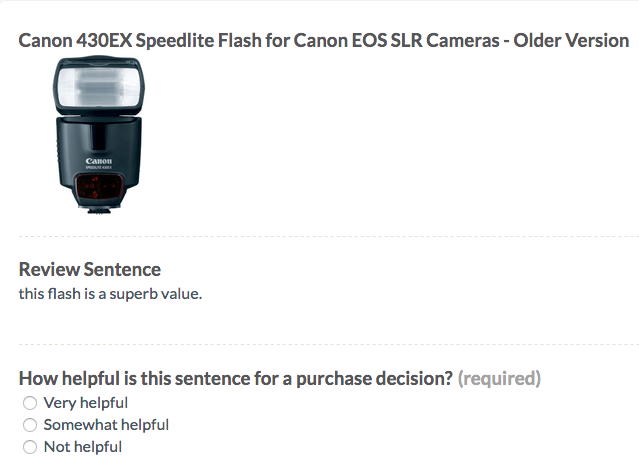}
	\caption{A single annotation as presented to the annotators.\label{fig:annotation}}
\end{figure}

Figure~\ref{fig:annotation} displays a single annotation task as presented to the annotators.

\paragraph{Helpfulness Distribution within the Train Set}

\begin{figure}[ht]
	\begin{tikzpicture}
		\begin{axis}[
			width=1\linewidth,
			height=5cm,
			legend style={at={(0.5,1)},anchor=south,draw=none},
			legend entries={Train,Random},
			legend columns={3},
			grid style=dashed,            
			no markers,
			very thick,
			xlabel=Helpfulness,
			ymajorgrids=true,
			ylabel=Frequency,
			]
			\addplot[orange,smooth, thick] table [] {data/train_hist.csv};
			\addplot[green, dotted, smooth, thick] table [] {data/random_hist.csv};
		\end{axis}
	\end{tikzpicture}
	\caption{\label{fig:help-dist}
		Helpfulness distribution among the train set.
		The green, dotted line represents the expected distribution if the annotators would have chosen an answer uniformly at random.
		The most frequent helpful score on the training set is 1.3.
	}
\end{figure}
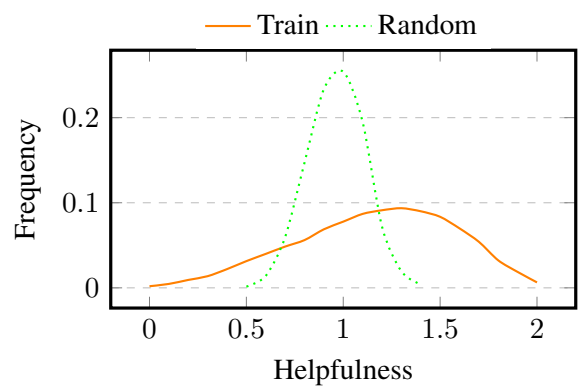

Figure~\ref{fig:help-dist} depicts the helpfulness distribution within the train set.

\section{Annotation Analysis}
\label{app:analysis_annotation}

Table~\ref{tab:contrast-data} provides some statistics for the helpful and unhelpful subsets.

\begin{table}[ht]   
	\centering
	\resizebox{\columnwidth}{!}{
		\begin{tabular}{@{}lcccc@{}} \toprule
			& \# Helpful & \# Reviews & \# Sentences & Mean \\
			& Votes & & & Sentences\\
			\cmidrule{2-5}
			Helpful Set   & $\ge$ 50  & 101   & 5032  & 49.82   \\
			Unhelpful Set     & 0         & 3803  & 22030 & 5.79    \\
			\bottomrule
	\end{tabular}}
	\caption{
		Contrasting Sentence Helpfulness Score with Review Helpfulness votes --- Helpful and unhelpful sets statistics.%
		\label{tab:contrast-data}
	}
\end{table}

\section{Analysis of Helpful Sentences}
\label{app:analysis_help}

\paragraph{Length}

Sometimes a sentence may be long but provide information that is not very helpful for customers. As an example for a long sentence with low helpfulness score, consider the sentence ``As engineers, you must learn to handle the technical aspects, the social and political gamesmanship, the public reactions to our work and the daunting challenge of staying at pace with rapid developments in our fields'' that was taken from a review about the book Mastery which deals with becoming a master in one's field. Indeed, this sentence is long and informative but it talks about engineers while the book itself is meant for everyone interested in mastery, not necessarily engineers.

\paragraph{Sentiment}

Consider the following sentence, demonstrating that sentiment is not always necessary for helpfulness: ``It teaches parents to use twelve key strategies in everyday scenarios to help them in providing experiences to promote healthy connections in the brain.'' This sentence was deemed helpful by the annotators but does not express any sentiment, it merely states a fact about the product.

\section {Surfacing Representative Helpful Sentences}
\label{app:surface}

\paragraph{Preprocessing}
First, HTML markup is removed from each review using the BeautifulSoup\footnote{\url{https://pypi.org/project/beautifulsoup4/}} package and then the cleaned texts are split into sentences using the Spacy\footnote{\url{https://spacy.io/}} package. 
Next, sentences with character length outside the range of $[30, 200]$ are removed. We chose these thresholds based on manual inspection under the assumption that extremely short sentences will not be very helpful for customers while extremely long sentences will result in a frustrating customer experience (especially in the scenario of voice interactions). In the movies domain, for example, a typical short sentence would state that \emph{``This movie was really good.''} or that \emph{``It's a must see film.''}, statements that do not contribute much on top of the star rating.
In our dataset, long sentences are quite rare while short sentences, on the other hand, are more common and form $10\%$ of the sentences.

\paragraph{Sentence Selection}
The Sentence Selection component is in charge of selecting a single sentence that is both helpful and well supported. We enforce a minimum support of $5$, as we observed that such a limit increases the overall quality of the sentence and avoids surfacing esoteric opinions.
After applying this threshold, we rank the remaining sentences according to the formula:
\begin{equation}
	\label{eq:final}
	\text{support} \times{} \text{helpful}^\alpha
\end{equation}
where $\alpha$ is a boosting parameter. 
To derive an appropriate value for $\alpha$ we conducted another annotation task in which annotators were asked again to score the helpfulness of the sentences presented to them. This time we consider all the sentences that are not dominated by any other sentence, i.e. we consider sentence $s$ if and only if there is no sentence $s'$ such that both $\text{helpful}(s') > \text{helpful}(s)$ and $\text{support}(s') > \text{support}(s)$, in other words, we asked to annotate all the sentences from the Pareto front with respect to helpfulness and support. 
Each sentence was joined with a prefix that quantifies its support as in \emph{20 customers agreed that: has very good pic quality and extremely easy to use}.
We optimized Formula~\ref{eq:final} by minimizing the Kullback–Leibler divergence between the score distribution (softmax) from the annotators and the score distribution from the formula (softmax) and obtained $\alpha=38.8$.
While this value may seem enormous at a first glance we note that the helpfulness score obtained from our model for the sentences in the Pareto set tend to be very close to each other while their support may vary considerably.
To put this number into proportion, consider two sentences with support and helpfulness $(20, 1.5)$ and $(10, 1.52)$ respectively, then the final score of the first sentence will only be slightly better than the final score of the second sentence (which is the expected behavior as its support is twice as large).

Interestingly, the experiment confirmed our hypothesis that customers perceive highly supported sentences as more helpful compared to the case when no support information is given.\footnote{We experimented with another prefix, that only states \emph{``one customer thought that''.}}

\section{Model Output Statistics} 
\label{app:stats}

As for end-to-end results on our $123$ products, our model found a positive helpful sentence for $114$ products, and a negative helpful sentence for $16$ products. The low coverage for negative helpful sentences might be explained by our choice to concentrate on popular products (having more than $100$ reviews) which are probably of better quality than random products.

 In Table~\ref{tab:ex} we present the selected sentence, its helpfulness score and its top supported sentences along with their similarity scores for $3$ products.

\begin{table*}
	\centering
	\resizebox{0.9\textwidth}{!}{
		\begin{tabular}{@{}lp{12cm}r@{}}
			\multirow{7}{24mm}{\centering\includegraphics[width=20mm,keepaspectratio]{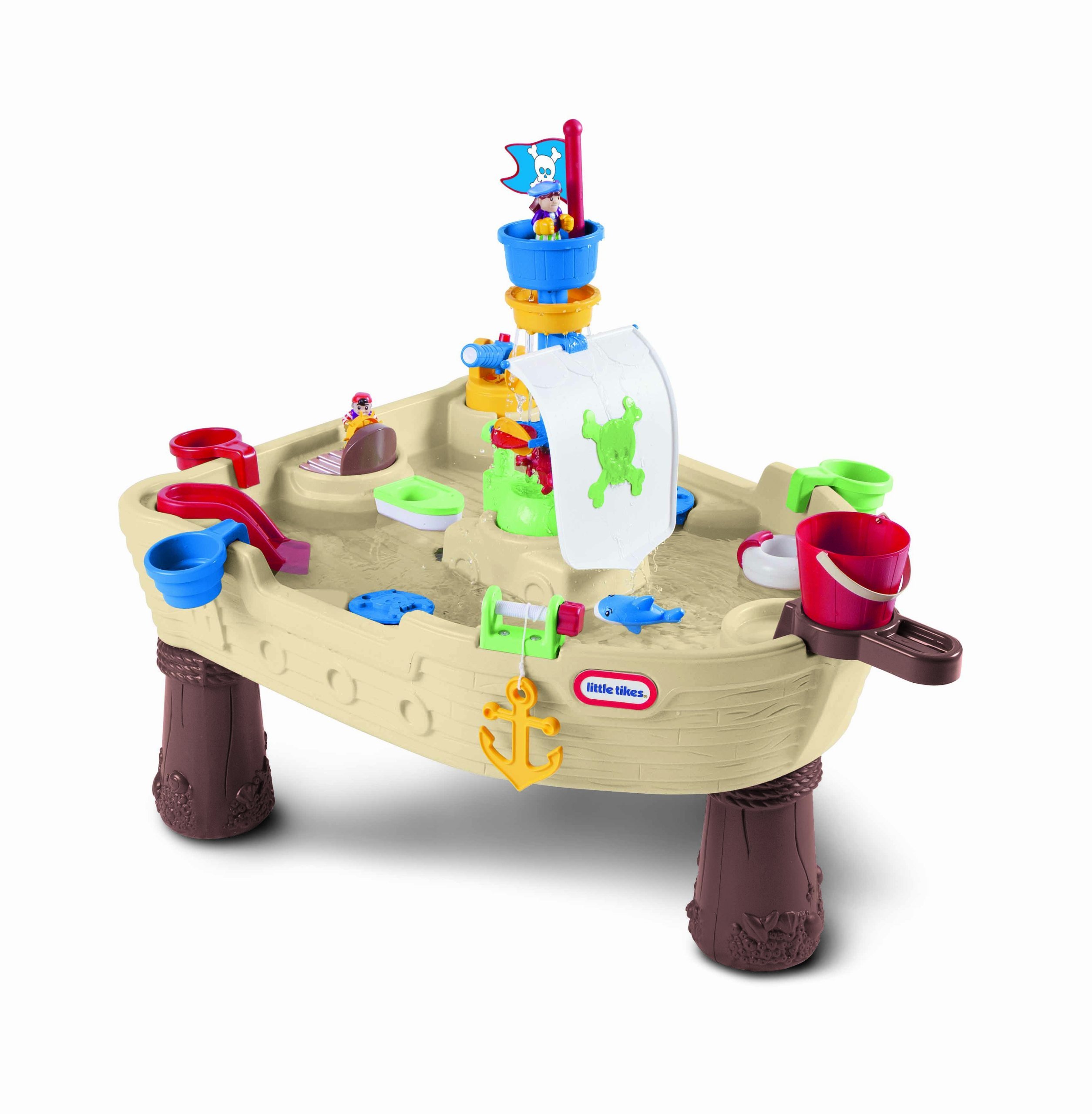}}
			& \textbf{It was easy to put together, is the perfect height, and very durable.} & 1.6\\
			\cmidrule{2-3}
			& It was easy to put together and is sturdy. & 0.94 \\
			& Sturdy and easy to put together. & 0.92 \\
			& Also, it was very easy to put together. & 0.92 \\
			& It's sturdy and cleans easily. & 0.91 \\
			& Pretty sturdy, too, and easy to handle. & 0.91 \\
			\midrule
			\multirow{7}{24mm}{\centering\includegraphics[width=20mm,keepaspectratio]{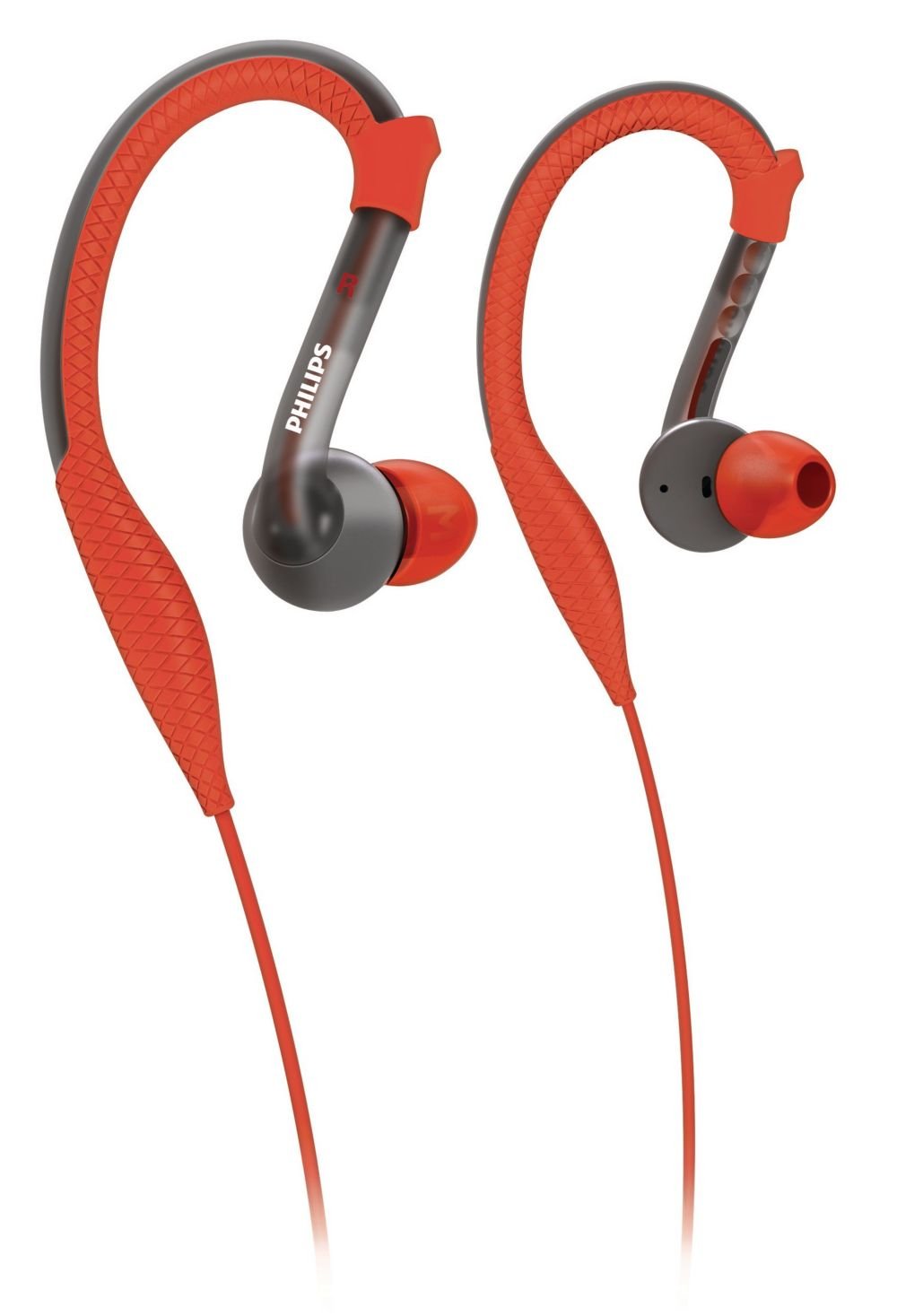}}
			& \textbf{They fit well and produce good sound quality.} & 1.7\\
			\cmidrule{2-3}
			& fits great and has great sound & 0.93 \\
			& The sound is pretty good as well. & 0.93 \\
			& Great Pair works great with excellent sound & 0.93 \\
			& They do have good sound, when I can keep them in. & 0.91 \\
			& These sound great and fit very well. & 0.9 \\
			\midrule
			\multirow{7}{24mm}{\centering\includegraphics[width=20mm,keepaspectratio]{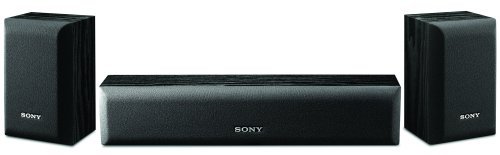}}
			& \textbf{Quality good, price okay, sound output great.} & 1.7\\
			\cmidrule{2-3}
			& For the price point, this set delivers good sound. & 0.92 \\
			& Very good sound for the cost!! & 0.92 \\
			& It works and the sound quality is good. & 0.9 \\
			& Good quality speakers that provide great sound. & 0.9 \\
			& For me ... the sound was very good. & 0.88 \\
			\midrule

	\end{tabular}}
			\caption{The selected sentence, its helpfulness score and its top supported sentences along with their similarity scores for $3$ products.}
		\label{tab:ex}
\end{table*}